%% file: iclr2020_conference.tex
\documentclass{article} 
\usepackage{iclr2020_conference,times}

\input{math_commands.tex}

\usepackage{hyperref}
\usepackage{url}
\usepackage{graphicx}
\usepackage{booktabs}

\title{Predicting Unplanned Readmissions \\ with Highly Unstructured Data}

\author{Constanza Fierro\thanks{Currently at Google (\texttt{constanzam@google.com}).}, Jorge Pérez \\
Department of Computer Science, Universidad de Chile \& \\
Millennium Institute for Foundational Research on Data\\
\texttt{\{cfierro,jperez\}@dcc.uchile.cl} \\
\AND
Javier Mora \\
Clinica Las Condes, Santiago, Chile \& \\
Surgery Department, School of Medicine, Universidad de Chile \\
\texttt{jmorap@clc.cl} \\
}

\iclrfinalcopy 
\begin{document}
\maketitle

\begin{abstract}
Deep learning techniques have been successfully applied to predict unplanned readmissions of patients in medical centers. 
The training data for these models is usually based on historical medical records that contain a significant amount of free-text from 
admission reports, referrals, exam notes, etc.
Most of the models proposed so far are tailored to English text data and assume that electronic medical records follow standards common in developed countries.
These two characteristics make them difficult to apply in developing countries that do not necessarily follow international standards for registering patient information, or that store text information in languages other than English.

In this paper we propose a deep learning architecture for predicting unplanned readmissions that consumes data that is significantly less structured compared with previous models in the literature. We use it to present the first results for this task in a large clinical dataset that mainly contains Spanish text data.
The dataset is composed of almost 10 years of records in a Chilean medical center. On this dataset, our model achieves results that are comparable to some of the most recent results obtained in US medical centers for the same task (0.76 AUROC).


\end{abstract}


\section{Introduction}
Unplanned hospital readmissions are frequent and expensive. 
The~\citet{centers2017hospital} showed that between 2013 and 2016, 15.5\% of the annual hospitalizations correspond to unplanned readmissions within 30 days after discharge. 
Another study showed that US unplanned re-hospitalizations in 2004 had a cost of \$17.6 billion~\citep{jencks2009rehospitalizations}. 
Moreover, unplanned readmissions are used as a measure of quality; 
To encourage an improvement in the quality of health services, the US and other countries have set penalties to early readmissions~\citep{kristensen2015roadmap}. 
For these reasons, it is expected that hospitals are interested in methods that can accurately predict patients that have a high risk of being readmitted, so they can focus on that group and conduct follow up calls, schedule recurrent visits or give more healthcare education.

Deep learning~\citep{DL} has emerged as one successful machine learning technique for predicting several health care measures. 
In particular, one of the best results reported so far for predicting unplanned readmissions was produced by~\citet{rajkomar2018scalable} with a deep learning architecture.
The most successful deep learning models use free-text as one of their main sources for training data, and most of the models have been trained for the English language. 
Moreover, besides text data, these models rely on Electronic Health Records (EHR) that follow standards such as the Fast Healthcare Interoperability Resources (FHIR) or the National Center for Biomedical Ontology (NCBO) BioPortal.
Thus, they cannot be directly applied to medical centers that do not necessarily follow international standards for registering health-care records, or in which text information is stored in languages other than English.
One possible partial solution would be to perform a data migration to a standard EHR format.
But for public hospitals in developing countries, such a solution is totally prohibited in terms of the amount of resources it may cost. 
This motivates the search for a more affordable solution.

In this paper we present a deep learning model that predicts unplanned readmissions consuming data that is considerably less structured compared with the models proposed in the literature. 
For training our model we use almost 10 years of patient records in a Chilean medical center.
In this dataset, a big part of the records are structured in ad-hoc formats designed by the medical center itself, and all the text data is in the Spanish language. 

Our preliminary results are encouraging: 
for the task of predicting unplanned re-admisions after 30 days of discharge, we achieve 0.76 AUROC (Area Under the Receiver Operating Characteristic curve). 
This result is comparable to the one by~\citet{rajkomar2018scalable} for two US medical centers for the same task in which they obtained 0.76 and 0.77 AUROC. It is also comparable to the AUROC obtained by~\citet{jamei2017predicting} of 0.78.
Our results outperform also the results by
the Length-Admission-Comorbidity-Emergency (LACE) method~\citep{van2010derivation} that has an estimated AUROC of 0.72~\citep{jamei2017predicting}.
To the best of our knowledge, our reported results are the first results for our task in a large clinical dataset that mainly contains Spanish text data.



\section{Related work}

Deepr~\citep{nguyen2017mathtt} is an end-to-end system based on a convolutional network~\citep{DL} that predicts future risk for patients. 
It is validated predicting unplanned readmissions after discharge. 
Deepcare~\citep{pham2016deepcare} is a model that predicts future medical outcomes based on a Long Short-Term Memory recurrent network (LSTM)~\citep{hochreiter1997long}. 
Its performance is tested for disease progression modeling and readmission prediction in diabetes.
Both Deepr and Deepcare are based on diagnosis codes and intervention (procedures and medications) codes. 
In Chile there's no regulation that forces hospitals to record intervention codes, so in our dataset these codes are not present.
Thus, we cannot map EHRs to feature vectors using this approach. 
Deep Patient~\citep{miotto2016deep} is an unsupervised deep feature learning method to derive a general-purpose patient representation from EHR. It is strongly based on the use of the NCBO BioPortal~\citep{musen2011national} to extract biomedical concepts from text in English, which we can certainly not use in Spanish. 

In 2018,  Rajkomar et al.~proposed a different way to represent the patient history using the entire raw healthcare records~\citep{rajkomar2018scalable}. They validated their approach in several tasks including unplanned readmissions. 
They assume that the data can easily be transformed to FHIR resources~\citep{mandel2016smart}. 
In our dataset there is a significant portion of the data that does not have a matching FHIR resource (see the {\bf Form} entity in Section \ref{sec:data}).

There are some recent works that obtain good AUROC performance by exploiting specific hand-crafted features that are then fed to neural network architectures~\citep{jamei2017predicting, lin2019analysis}. 
In our work we explicitly avoid the manual creation  of features, as most of our data is highly unstructured and stored as raw text. 


\begin{table}
    \centering\small 
    \caption{Number of data points for each (sub)entity in CLC data.}
    \label{table:numbers}
    \begin{tabular}{cccccccc}
    \toprule
    \textbf{Patient} & \textbf{Problem} & \textbf{Encounter} & \textbf{Diagnosis} & \textbf{Order} & \textbf{Form} & \textbf{Clinical note}  \\ \midrule
     1,475,224 & 943,024 & 7,955,974 & 6,188,676 & 19,937,072 & 21,508,633 & 18,448,316 \\
     \bottomrule
    \end{tabular}
\end{table}

\section{Data}\label{sec:data}

We use the historical data of Clinica Las Condes (CLC) of Chile between November 2009 and December 2018. 
To use this data we first obtained the approval of the respective  ethics committee, and it was conditioned on the no use of personal traceable data. 
The EHR in Clinica Las Condes includes: free text of reason of visit, diagnosis information, clinical notes, medication orders, procedure orders, and socio-demographic variables. 
This data was stored in relational databases and modelled with the following three main \emph{entities}: (1)~\textbf{Personal} data, (2)~\textbf{Problem} and (3)~\textbf{Encounter}. Each encounter (which corresponds to a one time visit) may contain several of the following sub-entities: (3.1)~\textbf{Diagnosis}, (3.2)~\textbf{Order}, (3.3)~\textbf{Form} and (3.4)~\textbf{Clinical Note}. 
Except for the personal data, a patient history may have zero or multiple instances of each entity. The numbers corresponding to each (sub)entity are shown in Table~\ref{table:numbers}.

The (sub)entities in our data can be described as follows: \textbf{Personal} data contains information about patiens such as birth date, age and nationality. The \textbf{Problem} entity describes chronic illnesses or medical conditions that last for a long period of time (e.g.: asthma). 
The \textbf{Encounter} entity contains one time visit information consisting of: beginning and end date, reason of the visit and type of the visit. As we said, each encounter can contain zero or several sub-entities. The \textbf{Diagnosis} sub-entity, describes the details of the diseases diagnosed in that encounter. The \textbf{Order} corresponds to medical exams orders or prescription drugs. The \textbf{Clinical Note} corresponds to free text for any type of information that a doctor wants to add. Finally, the \textbf{Form} sub-entity, which is the most numerous in our dataset (21.5M data points) was created at CLC as a very flexible way of storing different types of information.  
The data in a \textbf{Form} entity is a list of key-value pairs, where each key and value are strings that may represent any type of data: free text, categorical classes, numbers, measurements, dates, etc. For example, there are forms to describe the patient state when he/she arrived at the emergency room, or to register the vital signs of an inpatient.



\begin{figure}[t]
\begin{center}
\includegraphics[width=0.75\textwidth]{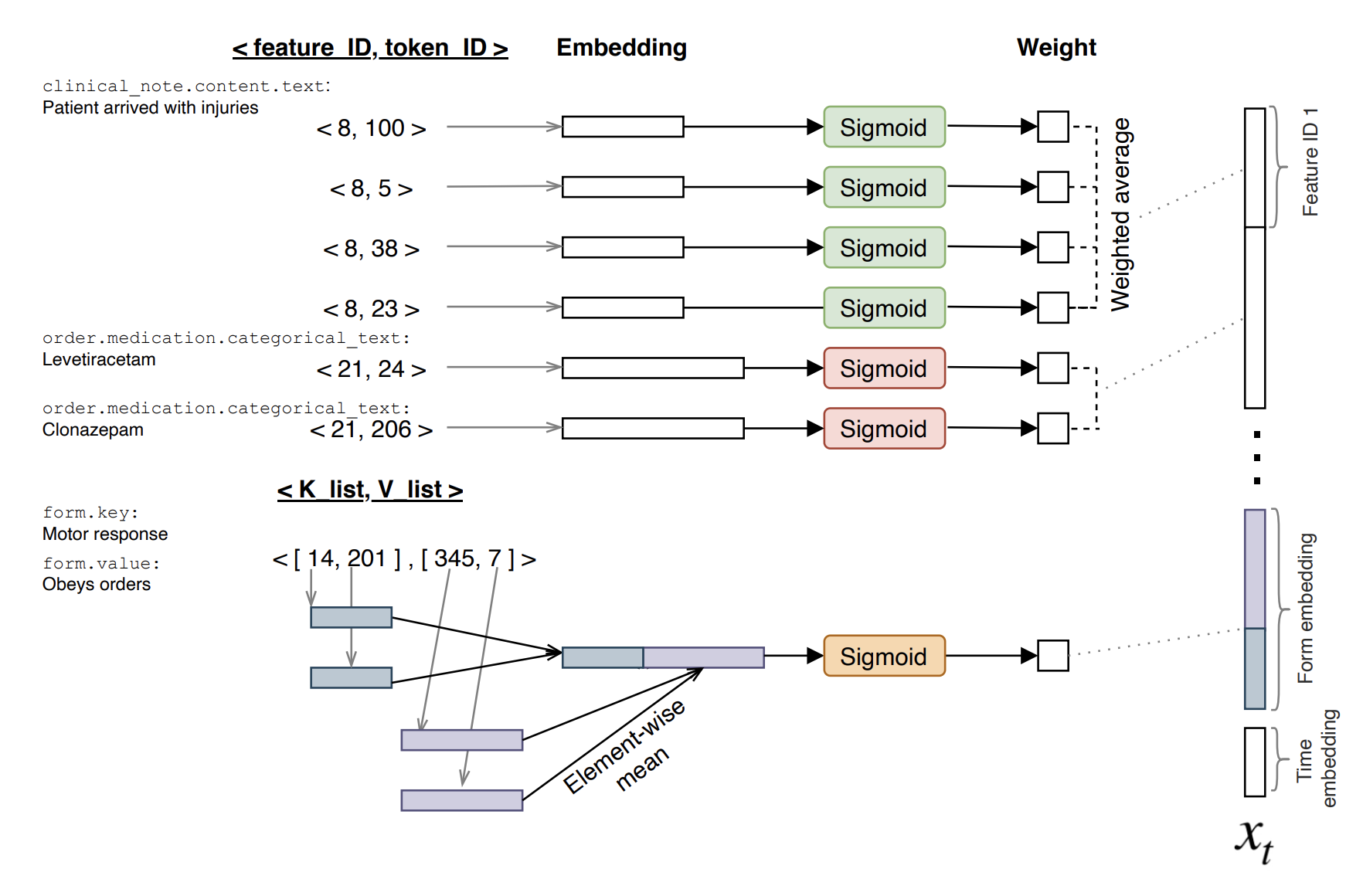}
\end{center}
\caption{Model architecture to create the vector representation within a time window.}
\label{fig:our_arq}
\end{figure}

\section{Methodology}


To define valid admissions for our task we followed as much as possible the definition of the CMS~\citep{unplanned_readmissions}, so we just consider admissions where: the patient didn't die, the patient wasn't discharged against medical advice, and the patient was not transfered to another hospital. 
We were not able to totally follow the CMS definition for unplanned readmission primarily because there was no treatment or procedures codes on the dataset. 
In cooperation with nurses and computer technicians of the hospital, we decided to use the ``pre-registration'' records as a proxy. 
Everytime an admission is scheduled in advance a pre-registration record is created.
So every admission that does not have a pre-regristraion was considered as unplanned.
We also considered as unplanned every admission that has a pre-registration during the 24 hours before the hospitalization is created.

With the definition of valid admission and unplanned readmission settled, we created our training dataset considering each example as all the available patient history until a valid admission.
We labeled this example as \textbf{1} if the patient has an unplanned re-admision after 30 days of discharge.
Otherwise we labeled that example as \textbf{0}.
Our cohort then has 6.18\% (11,459) positive examples, and 93.82\% (173,864) negative examples. It is important to note that each example corresponds to one valid admission, so one patient that has more than one valid hospitalization can produce multiple examples. Inspired by~\citet{arango2019hate}, we split the train (80\%), validation (10\%) and test (10\%) sets by patient and not by example.
Since this was made randomly, the overall rates of positive/negatives examples were preserved in each set, as we can see in Table \ref{table:distribution_sets}.

\subsection{EHR representation}\label{subsec:representation}
As we said in the introduction, we cannot completely rely on EHR representations used by the models proposed so far in the literature. 
So we decided to follow an idea similar to~\citet{rajkomar2018scalable} in order to use all types of data (text, codes, names, etc.).
We consider each entity data and each of their subentities and labels as \textbf{features}.
For example in our dataset we have the feature ``\texttt{order.medication.categorical\_text}''.
Each of these features can be associated with several different \textbf{tokens}. 
For example, for the above mentioned feature we can have  ``\texttt{levetiracetam}'', ``\texttt{clonazepam}'' and ``\texttt{enalapril}'' as possible tokens.
For the case of free-text data, we follow a very simple approach.
As an example, consider the feature ``\texttt{clinical\_note.content.text}''
and assume that it is associated with the text ``\texttt{patient arrived with injuries}''.
In this case we just consider it as four pieces of data for each of the words in the text. That is, the feature is associated with the tokens ``\texttt{patient}'', ``\texttt{arrived}'' ``\texttt{with}'', and ``\texttt{injuries}''.
After this process we can represent each piece of atomic data as
a pair \texttt{(feature\_ID, token\_ID)}.

Given that Forms are really ad-hoc resources, we use a different representation for them.
As we have said, Forms are just a set of key-value pairs.
Since both keys and values can be defined with arbitrary text, we use sequences of token ids to represent both.
Thus, a key-value pair is represented as \texttt{(K\_list, V\_list)}.
At the end, we have two types of data representation, one as pairs \texttt{(feature\_ID, token\_ID)} to represent non-Form data, and another as pairs \texttt{(K\_list, V\_list)} to represent Form data.
Finally, since every data can be associated with a timestamp, we can represent the patient history as a list of time-ordered set of pairs. 


\begin{table}[]
\centering\small
\caption{Labels distribution by set.}
\label{table:distribution_sets}
\begin{tabular}{lcccc}
\toprule
\textbf{} & \textbf{\begin{tabular}[c]{@{}c@{}}Number \\ of examples\end{tabular}} & \textbf{\begin{tabular}[c]{@{}c@{}}Percentage\\ of the total\end{tabular}} & \textbf{\begin{tabular}[c]{@{}c@{}}Percentage of\\ positives examples\end{tabular}} & \textbf{\begin{tabular}[c]{@{}c@{}}Percentage of\\ negative examples\end{tabular}} \\ \midrule
Total & 185.323 & -- & 6.18\% & 93.81\% \\
Train & 148.510 & 80.13\% & 6.23\% & 93.76\% \\
Validation & \phantom{0}18.318 & \phantom{0}9.88\% & 5.81\% & 94.18\% \\
Test & \phantom{0}18.495 & \phantom{0}9.97\% & 6.11\% & 93.88\% \\ \bottomrule
\end{tabular}
\end{table}

\subsection{Model}\label{subsec:model}
We use embeddings to represent each feature and token. Following~\citet{rajkomar2018scalable} we
consider 12 hours windows. For each feature we calculate a
weighted average between all the token embeddings for that feature that occur in the time window.
In our case we computed the weights for each token by using a neural network (NN) with one hidden layer and a single output neuron with a sigmoid activation. We have one NN for each feature, and each NN receives as input the embedding of a token (upper-left in Figure~\ref{fig:our_arq}).

We consider Form embeddings but computed in a different way. 
Recall that each key-value pair in a Form is represented as a pair of lists of tokens \texttt{(K\_list, V\_list)}.
For each \texttt{(K\_list, V\_list)} pair occurring in the time window, we first average all the token embeddings of \texttt{K\_list} to obtain an embedding $k$, and separately we averaged the token embeddings of \texttt{V\_list} to obtain the embedding $v$. 
We then use the concatenation $[k;v]$ as input for a NN (one hidden layer, sigmoid output) to compute a weight.
Finally, we use these weights to compute a weighted average of all the $[k;v]$ vectors representing every key-value pair in the Forms of the time window (right side in Figure~\ref{fig:our_arq}).

To obtain the representation of the time window, we just concatenate the embeddings representing each feature and the embedding representing the Forms, plus an embedding that represents a relative time w.r.t.~the time of discharge ($x_t$ in Figure~\ref{fig:our_arq}).
Finally, the complete history is a sequence of those 12 hours window vectors.

To make a prediction, we use the above described representation of the history of a patient to feed a LSTM.
We use the last hidden state of the LSTM to perform a logistic regression to predict the probability for each class (1 and 0).
We optimized everything jointly (LSTM, embeddings and weighted averages) to minimize the log-loss of the logistic regression using the Adam optimizer~\citep{kingma2014adam}. 
We used embedding dropout and LSTM hidden dropout.

\paragraph{Baseline model}
As in others studies that predict unplanned readmissions~\citep{nguyen2017mathtt, huang2019clinicalbert} we  compared our model to a Bag of Words (BOW) with Term-Frequency Inverse-Document Frequency (TF-IDF) and Logistic Regression (LR) output. 
We used the EHR representation (Section \ref{subsec:representation}) but we selected the 5,000 more relevant tokens according to its TF-IDF value. 
Each token TF-IDF value was calculated as the average of its TF-IDF value in each document, and we use each patient timeline as a document. 
\vspace*{-5pt}

\section{Results}
We trained the LSTM model for 6 epochs. 
The baseline was trained for 8 epochs.
On each training we test the model after one epoch on the validation set and we chose the one that performed the best according to the AUROC value. 
Tables \ref{table:results} and \ref{table:precision_recall_test}  show the performance of the chosen models on the test set. 
We also tried to do oversampling of the positive values till we reach 20\% on the training set but it obtained worse results compared with no oversampling.
Although our AUROC (0.76) is comparable with that of similar work~\citep{rajkomar2018scalable}, our recall is significantly low. We note that~\citet{rajkomar2018scalable} did not report precision nor recall, so we cannot compare with that work on those metrics. On the other hand, the AUROC achieved by our model is above the LACE approximated AUROC performance of 0.72~\citep{jamei2017predicting}. 
The precision and recall of LACE are 0.21 and 0.49, respectively~\citep{jamei2017predicting}.
Improving the recall of our method is part of our future work.



\begin{table}[t!]
\parbox{.4\linewidth}{
\centering\small
\caption{AUROC results}
\label{table:results}
\begin{tabular}{lc} \toprule
 \textbf{Model} & \textbf{AUROC} \\ \midrule
Baseline & 0.5807 \\ \addlinespace
LSTM  & \textbf{0.7626} \\
\, +oversamp. & 0.7499 \\
\bottomrule
\end{tabular}}
\parbox{.5\linewidth}{
\centering\small
\caption{Precision and recall by class}
\label{table:precision_recall_test}
\begin{tabular}{lcccc}\toprule
 \textbf{Model} & \textbf{Prec. 0} & \textbf{Rec. 0} & \textbf{Prec. 1} & \textbf{Rec. 1} \\ \midrule 
Baseline & \textbf{0.95} & 0.96 & 0.26 & 0.23 \\ \addlinespace
LSTM  & \textbf{0.95} & \textbf{0.99} & \textbf{0.52} & 0.13 \\
\, +oversamp. & \textbf{0.95} & 0.97 & 0.36 & \textbf{0.29} \\ \bottomrule
\end{tabular}}
\vspace*{-10pt}
\end{table}

\vspace*{-5pt}

\section{Conclusion}
Our proposed model achieves good results on the Chilean clinical dataset, and it presents the first experience of its kind on Spanish data.
Our trained model can be directly used to deploy a system to be used by CLC.
We are currently working with the CLC professionals on reaching that goal.

As future work we would like to visualize the weights used for averaging in every time window to try to understand where the model is focusing to make the predictions.
We would also like to improve the recall of our model possibly by modifying the loss function to weight more on the minority class. 
This should be done taking into account the hospital goals and resources, because it would probably imply to have more false positives. 

\subsection*{Acknowledgments}

We thank all the medical and technical personnel at Clinica Las Condes that helped us complete the project. 
Special thanks to Marcelo Arias, head of the CLC Informatics Department, whose support was crucial to have access to the data.
This work was supported by Clinica Las Condes under the Academic Research and Development program (PIDA 2018-002), and by the Millennium Institute for Foundational Research on Data.

\bibliography{iclr2020_conference}
\bibliographystyle{iclr2020_conference}

\end{document}

%% file: math_commands.tex
\usepackage{amsmath,amsfonts,bm}

\def\eqref#1{equation~\ref{#1}}

\def\1{\bm{1}}

\DeclareMathAlphabet{\mathsfit}{\encodingdefault}{\sfdefault}{m}{sl}
\SetMathAlphabet{\mathsfit}{bold}{\encodingdefault}{\sfdefault}{bx}{n}

